\documentclass[conference]{IEEEtran}

\usepackage{amsmath}
\usepackage{amssymb}
\usepackage{booktabs}
\usepackage{cite}
\usepackage{graphicx}
\usepackage{multirow}
\usepackage{tikz}
\usepackage{url}
\usetikzlibrary{arrows.meta,positioning,shapes.geometric}

\newcommand{\EOneReplicas}{2}
\newcommand{\EOneTransitions}{6400000}
\newcommand{\EOneViolations}{0}
\newcommand{\EOneAblations}{10}
\newcommand{\ETwoGroups}{542}
\newcommand{\ETwoAttempts}{2768}
\newcommand{\ETwoRetainedGroups}{122}
\newcommand{\ETwoViolations}{0}
\newcommand{\ETwoUpper}{0.0243}
\newcommand{\ETwoLate}{2}
\newcommand{\ETwoWeakOvershoot}{280}
\newcommand{\EThreeCells}{895}
\newcommand{\EThreeStrictObservations}{155}
\newcommand{\EThreeStrictCalls}{0}
\newcommand{\EThreeDivergences}{150}
\newcommand{\EThreeTamperRejections}{60}
\newcommand{\EThreeUnsafeAliases}{930}
\newcommand{\EThreeCustody}{310}
\newcommand{\EFourCells}{132}
\newcommand{\EFourProcesses}{660}
\newcommand{\EFourMaxSeconds}{408.47}
\newcommand{\EFourMaxRSSMiB}{968.95}
\newcommand{\EFourInterningRatio}{0.349}
\newcommand{\ESixMutations}{1470}
\newcommand{\ESixDetections}{1470}
\newcommand{\ESixControls}{120}
\newcommand{\ESixFalseAlarms}{0}
\newcommand{\ESixRepresentation}{30}
\newcommand{\ESixConstraint}{90}
\newcommand{\ESixInapplicable}{90}
\newcommand{\ESevenPairs}{360}
\newcommand{\ESevenTrees}{720}
\newcommand{\ESevenFullCaptured}{360}
\newcommand{\ESevenFullDeclared}{360}
\newcommand{\ESevenModelOnlyMissing}{120}
\newcommand{\ESevenToolOnlyMissing}{240}

\newcommand{\EHostedACompleted}{34}
\newcommand{\EHostedAErrors}{128}
\newcommand{\EHostedBCompleted}{16}
\newcommand{\EHostedBErrors}{4}
\newcommand{\EHostedConservativeSpend}{4.35704260}
\newcommand{\EHostedRemainingGuard}{2.64295740}

\newcommand{\EHostedATrials}{162}

\newcommand{\EHostedAUtilityDelta}{-0.0123}
\newcommand{\EHostedAUtilityCILow}{-0.0385}
\newcommand{\EHostedAUtilityCIHigh}{0.0000}

\newcommand{\EHostedBTrials}{20}
\newcommand{\EHostedBPairs}{10}

\title{Pollard: Governed Semantic Execution Trees for Framework-Independent AI Agent Control and Replay}

\author{\IEEEauthorblockN{Muntaser Syed}\IEEEauthorblockA{Florida Institute of Technology\\Melbourne, Florida, USA}\and\IEEEauthorblockN{Marius C.\ Silaghi}\IEEEauthorblockA{Florida Institute of Technology\\Melbourne, Florida, USA}}

\begin{document}

\maketitle

\begin{abstract}
Agent frameworks can persist checkpoints, trace tool calls, and limit resource use, but these facilities often assign different identities to execution, governance, and replay. We present Pollard, a framework-independent runtime based on governed semantic execution trees. A step identity commits to canonical input, kind, attempt, and parent identity; the same tree records completion, refusal, resource settlement, and branch history. Strict replay returns a verified stored result without invoking the wrapped operation, while a transactional store coordinates reservations across concurrent workers. Under a signed deterministic freeze, two state-model replicas each executed \EOneTransitions{} transitions without an invariant violation, \ETwoViolations{} of \ETwoRetainedGroups{} retained-lease PostgreSQL groups violated an exact limit, strict replay made \EThreeStrictCalls{} external calls in \EThreeStrictObservations{} observations, and all \ESixDetections{} semantic mutations were detected with \ESixFalseAlarms{} alarms on \ESixControls{} controls. A \ESevenPairs{}-pair framework matrix preserved reference behavior and verified \ESevenTrees{} trees. A separate cost-capped hosted study exercised live provider paths, and every completed Pollard trial produced a verified tree. Provider compatibility and the frozen budget boundary limited the hosted utility comparison, so it is reported as integration evidence rather than benchmark-wide equivalence. Together, the studies support the core mechanisms under explicit lease, boundary, and seal-custody assumptions.
\end{abstract}

\section{Introduction}

Tool-using language model agents perform work through a sequence of model calls, retrievals, tool invocations, and external side effects. A final answer does not reveal which resources were consumed, why a tool call was permitted, or whether a later reconstruction repeated the same execution. Recent work documents budget overruns in deployed agent systems \cite{khan2026budgets}, proposes formal resource contracts \cite{ye2026contracts}, and identifies execution provenance as a basis for process-level accountability \cite{wang2026traces}. These lines of work expose a common systems problem: control decisions and execution evidence need a shared identity.

Existing systems address parts of this problem. Runtime policy languages mediate actions before execution \cite{wang2025agentspec}. Agent workbenches record tool responses and inject repeatable faults \cite{mazumder2026agentcheck}. Graph runtimes persist checkpoints and support forks, although resumed nodes may invoke models and tools again \cite{langchain2026timetravel}. Tool caches reduce repeated calls by keying requests from tool names and arguments \cite{zhai2026toolcaching}. None of these mechanisms, by itself, establishes that a stored result, a resource reservation, and an authorization decision refer to one ancestry-bound semantic step.

Pollard introduces a runtime primitive called a governed semantic execution tree. A caller retains its own control loop and wraps selected model or tool operations as semantic steps. Before a step runs, Pollard derives an identity from the step input and its parent, evaluates resource and action gates, and records the decision. After execution, it settles measured resource use and stores the result. In strict replay mode, a matching step reads the recorded result and does not call the wrapped operation.

This paper makes four contributions:

\begin{itemize}
    \item A formal execution model that binds semantic identity to causal ancestry while leaving orchestration with the caller.
    \item Admission and settlement rules for multi-dimensional resources, including exact safety conditions and an explicit bound for estimation error.
    \item A runtime implementation with in-memory, SQLite, and PostgreSQL stores, action registration, refusal nodes, branching, replay, sealing, and telemetry adapters.
    \item A signed evaluation of adversarial concurrency, replay nonexecution, causal separation from argument-only caching, integrity, storage, framework boundaries, and cost-capped hosted integration.
\end{itemize}

The central question is not whether content addressing, checkpoints, or resource counters are individually new. The question is whether one minimal execution identity can make governance and replay mutually checkable without requiring a new agent framework.

\section{Problem Setting}

\subsection{Execution provenance}

Provenance systems represent how entities and activities contribute to a result. The W3C PROV data model supplies a domain independent vocabulary for entities, activities, agents, and derivations \cite{moreau2013prov}. Scientific workflow systems demonstrate that automatic provenance capture and durable workflow execution can support reproducibility at scale \cite{huber2020aiida}. Agent execution adds two complications. First, model and tool calls may be nondeterministic. Second, a runtime may need to refuse an action before the action produces an output.

Pollard records operational provenance. It does not infer claim level evidence, semantic justification, or factual support. This boundary matters because agent provenance research distinguishes an execution graph from evidence relations that support a claim \cite{wang2026traces}.

\subsection{Admission before execution}

A post hoc trace can explain a call after it has happened, but it cannot prevent a known violation. Pollard therefore performs admission before invoking a wrapped operation. Admission considers estimated resource use, action registration, policy, schema validation, and confirmation. A denied attempt becomes a refusal node. The record therefore contains alternatives that were considered but not executed.

An estimate can differ from measured use. The system does not claim that every unknown future cost can be capped exactly. Instead, it separates exact meters from approximate meters and reports the assumptions needed for each guarantee.

\subsection{Replay semantics}

Replay has several meanings. A checkpoint runtime can resume from saved state and reexecute later nodes. A transport recorder can return a prior HTTP response. A tool cache can reuse a result for matching arguments. Pollard defines strict replay at the semantic step boundary: if the derived node identity and stored result match, the runtime returns that result and does not invoke the wrapped operation. This definition is testable with a sentinel that fails if any external operation is called.

\subsection{Trust and threat boundary}

The design addresses four classes of failure. First, concurrent workers may race during admission or repeat reserve and settle requests after losing an acknowledgment. Second, a process or database connection may fail between admission, external execution, settlement, and node persistence. Third, replay may find a node under the wrong ancestry or a stored result whose digest no longer matches. Fourth, an operator or storage fault may alter node fields, results, membership, or ordering. The evaluation injects each class at a declared boundary.

Pollard trusts its canonicalizer, hash implementation, runtime code, and the transactional semantics of the selected store. Exact shared-budget safety additionally trusts a completing owner to retain its reservation through settlement. Tamper evidence trusts an externally held seal digest when store deletion or replacement is in scope. These assumptions are explicit because no library can preserve a global cap during an unavailable database while also admitting new work, and a seal kept only beside the protected data cannot expose wholesale replacement.

The caller chooses which operations to wrap. Pollard cannot govern a model or tool invocation that bypasses that boundary, cannot validate the truth of a model response, and cannot reconstruct external world state from an old tool result. Consequently, integration coverage is measured against an out-of-band boundary proxy, and replay claims concern nonexecution and stored-result integrity rather than renewed correctness of an external service.

\section{Governed Semantic Execution Trees}

\subsection{Tree and identity}

Let an execution tree be $T=(V,E)$ with a distinguished root. Every non-root node has exactly one parent already in $V$, so an edge records causal placement rather than merely data similarity. A node is
\begin{equation}
n = (p, k, a, x, y, m),
\end{equation}
where $p$ is the parent identifier, $k$ is the step kind, $a$ is the attempt number, $x$ is the identity input, $y$ is an optional stored result, and $m$ is metadata. Completion and refusal are represented by stored node kinds and results; an operation that raises before producing a result is not converted into a completed node. Let $C_I$ be deterministic identity serialization and $d_I$ an identity domain separator. The semantic identifier is
\begin{equation}
I(n) = H(d_I \mathbin\Vert C_I(p,k,a,x)).
\end{equation}
The canonical identity domain contains null, booleans, integers, strings, lists, and string-keyed objects. Floating point values and raw bytes are rejected because platform-dependent encodings would weaken identity stability. Metadata and the stored result do not determine $I(n)$. A result is serialized once by a separate stable JSON serializer $C_R$ and committed as $D(n)=H(d_R\mathbin\Vert C_R(y))$.

The current cursor identifies the parent for the next step. Branching copies a cursor without copying nodes; rollback moves a cursor to an ancestor; pruning marks metadata for later garbage collection. These operations change reachability or future ancestry, not an existing node identity. Identical semantic inputs under the same parent converge on one identifier. Identical inputs under different parents remain distinct unless the hash function collides. This choice gives up some argument-only cache reuse to prevent a match from erasing causal history.

\subsection{Admission and settlement transition}

For scope $s$, let $B_s$ be settled use, $L_s$ its componentwise limit, and $R_s$ the active reservations. Reservation $j$ carries estimate vector $e_j$, owner identity, and expiry. A new step $i$ is admitted only if the store atomically establishes
\begin{equation}
B_s + \sum_{j\in R_s} e_j + e_i \leq L_s
\end{equation}
for every applicable scope. The same transaction creates $i$ in $R_s$. The wrapped operation then either raises, or returns $y_i$ with measured charge $c_i$. A successful settlement atomically removes $e_i$ and adds $c_i$ to $B_s$. A call failure releases the reservation and propagates the exception. A gate failure appends a refusal node containing the blocked kind and a digest of the blocked payload, then returns control by raising without invoking the operation.

Reservations are leases because a worker can disappear after admission. A lease may expire and make its estimate available to another worker. Therefore exact safety applies to calls whose owners retain the lease until settlement. If an expired owner later completes, its result is evidence of work performed outside that premise and is reported separately rather than folded into the zero-violation contrast.

\subsection{Replay and sealing}

Record mode executes and stores a missing step. Hybrid mode returns a stored result on a matching identity hit and otherwise records. Strict replay requires the node and result to exist and verify; a miss or divergence fails without calling the supplied operation. Verification recomputes $I(n)$, checks parent existence and traversal order, and recomputes $D(n)$ from the stored result text.

For deterministic preorder $(n_0,\ldots,n_t)$, Pollard forms a rolling seal. Let $z_{-1}$ be empty. Each entry is
\begin{equation}
z_i=H(d_S\mathbin\Vert C_I(i,I(n_i),p_i,k_i,D(n_i),z_{i-1})).
\end{equation}
The report publishes $z_t$ and the ordered entries. The index and previous digest expose reordering or truncation; node and result digests expose semantic changes. Detection of total store replacement still requires $z_t$ to be held outside the store.

\subsection{Guarantees}

\textbf{Lemma 1 (Ancestry separation).} If two nodes differ in $p$, $k$, $a$, or canonical $x$, then their identifiers differ unless $H$ collides. A changed ancestor therefore changes each descendant identifier along that path.

\emph{Proof sketch.} The differing field changes the domain-separated input to $H$. Each child commits to its parent's identifier, so induction extends the difference to every descendant on the path.

\textbf{Theorem 1 (Exact retained-lease safety).} Suppose reserve and settle are serializable, each completing owner retains its reservation, and $c_i=e_i$ for every admitted step. Then $B_s+\sum_{j\in R_s}e_j\leq L_s$ is invariant for every scope $s$, and settled use cannot exceed $L_s$.

\emph{Proof sketch.} The empty state satisfies the invariant. Reserve preserves it by the admission predicate. Exact settlement replaces $e_i$ with the equal $c_i$ and release only removes an estimate. Serializable transitions therefore preserve the invariant by induction. Because all active estimates are nonnegative, $B_s\leq L_s$.

\textbf{Corollary 1 (Estimated-meter bound).} Under the same transactional and lease assumptions, let $\epsilon_i=\max(0,c_i-e_i)$. Observed overshoot is at most $\sum_i\epsilon_i$ over the relevant settlement interval. The bound does not turn an estimated meter into an exact cap; it identifies the quantity that calibration must control.

\textbf{Proposition 1 (Replay nonexecution).} If strict replay returns, the returned result belongs to the requested identity, its digest verifies, and the caller-supplied operation was not invoked. A sentinel operation that always fails makes the last condition directly testable.

\textbf{Proposition 2 (Rolling-seal sensitivity).} Under collision resistance and an externally retained final digest, any change to a sealed node identity, parent, kind, stored result, membership, or deterministic order changes the final seal. Metadata absent from the seal remains outside this claim.

Table~\ref{tab:guarantees} separates each guarantee from its required premise and its observable boundary case. This separation is central to the evaluation: a violated premise is retained as a result, not relabeled as evidence for the narrower theorem.

\begin{table*}[t]
\caption{Guarantees, required premises, and explicitly excluded outcomes.}
\label{tab:guarantees}
\centering
\small
\setlength{\tabcolsep}{4pt}
\begin{tabular}{p{0.19\textwidth}p{0.38\textwidth}p{0.36\textwidth}}
\toprule
Guarantee & Required premises & Boundary outside the guarantee \\
\midrule
Ancestry separation & Stable canonicalization, domain separation, and collision-resistant hashing & Caller omits a causal input or invokes work outside the selected boundary \\
Exact resource safety & Serializable reserve-settle, exact estimates, and lease retained through completion & Estimate error, expired owner completion, or unavailable shared store \\
Strict replay nonexecution & Exact identity hit, stored result present, ancestry and digest verification & Reexecution mode or unwrapped external side effects \\
Refusal soundness & Every governed action passes the gate before invocation & Caller bypasses the runtime or policy cannot classify the action \\
Rolling tamper evidence & Deterministic traversal, protected hash implementation, externally retained final digest & Mutable metadata excluded from the seal or loss of external custody \\
\bottomrule
\end{tabular}
\end{table*}

\section{System Design}

\begin{figure}[t]
    \centering
    \resizebox{\columnwidth}{!}{\begin{tikzpicture}[
    node distance=4mm and 5mm,
    box/.style={draw, rounded corners, align=center, minimum height=6mm, text width=19mm, font=\scriptsize},
    store/.style={draw, cylinder, shape border rotate=90, aspect=0.22, align=center, minimum height=9mm, text width=16mm, font=\scriptsize},
    flow/.style={-{Latex[length=1.6mm]}, thin}
]
\node[box] (caller) {Caller owned\\agent loop};
\node[box, right=of caller] (identity) {Canonicalize\\and identify};
\node[box, right=of identity] (gates) {Budget and\\action gates};
\node[box, right=of gates] (operation) {Model or tool\\operation};
\node[store, below=8mm of gates] (store) {Tree\\store};

\draw[flow] (caller) -- (identity);
\draw[flow] (identity) -- (gates);
\draw[flow] (gates) -- node[above, font=\tiny] {admit} (operation);
\draw[flow] (gates) -- node[left, font=\tiny] {refuse} (store);
\draw[flow] (operation) |- node[pos=0.7, right, font=\tiny] {settle} (store);
\draw[flow] (store) -| node[pos=0.25, below, font=\tiny] {strict replay} (caller);
\end{tikzpicture}}
    \caption{Pollard mediates semantic steps without owning the agent loop. The replay path returns a stored result without invoking the wrapped operation.}
    \label{fig:architecture}
\end{figure}
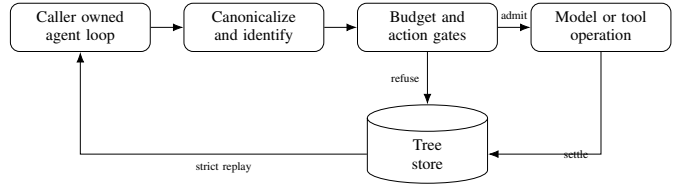

Figure~\ref{fig:architecture} shows the execution path. The caller owns planning and control flow. Pollard receives a semantic step request, derives its node identity, and consults resource and action gates. A denied step is recorded immediately. An admitted step either reads an existing result in replay mode or invokes the caller supplied operation. The store then settles resource use and persists the outcome.

\subsection{Boundary and execution lifecycle}

Pollard is a Python library rather than an agent scheduler. The caller opens a run, selects the parent cursor, and wraps a model call, tool call, or note. This placement is deliberate: only the application knows whether a framework node, provider request, retry, or tool callback is the semantic unit that should share identity with policy and replay. Synchronous and asynchronous runtimes implement the same lifecycle.

Before admission, the runtime validates the identity payload and computes the prospective node identifier. Record mode proceeds to gates and execution. Hybrid mode uses an identity match that has a stored result and records on a miss. Strict replay requires an existing result and verifies the stored path before returning it; any mismatch fails closed. Avoided charges remain mutable reporting metadata outside semantic identity.

On a record path, precheck estimates are collected from every meter and applied to each nested budget scope. Structural gates then validate a registered tool's name, version, schema, policy, and confirmation. A refusal is persisted before the corresponding exception is raised. Only after all gates pass does the runtime reserve shared capacity and invoke the caller function. Successful execution stops measurement, settles the reservation, constructs the result-bearing node, and advances the cursor. A function exception releases the reservation and propagates without fabricating a successful node.

This ordering does not claim exactly-once external side effects. If a provider or tool completes but the database becomes unavailable before settlement and node persistence finish, Pollard cannot safely infer whether repeating the external action is acceptable. The database operation is retryable; the external action remains an application-level ambiguity.

\subsection{Stores and concurrency}

The store interface separates node persistence from coordination. The memory store is intended for local execution. HashRope supplies append-oriented persistence with content interning. SQLite provides durable single-host storage and local arbitration. PostgreSQL supplies transactional reservations for concurrent processes and hosts. Store insertion is idempotent for an identical identity tuple and rejects an identifier conflict.

PostgreSQL reserve and settle requests carry a unique reservation identifier plus canonical request or charge digests. After connection loss, the client reconnects once and repeats the transaction. A committed reserve is recovered by the same identifier; a duplicate settle must carry the same charge digest and does not debit twice. If the retry cannot distinguish a committed operation from a failed one, the API raises an explicit uncertainty exception instead of guessing. This protects database accounting, but does not erase ambiguity about an external operation that ran between database failures.

A running call renews its lease at a fixed cadence derived from one third of the duration and bounded between 10 ms and 30 s. The runtime tracks the last confirmed deadline. If renewal is not confirmed before then, a completed result is settled and stored when possible, annotated with lease loss, and reported through an exception carrying its reservation and node identifiers. Its charge is excluded from retained-lease claims. Operators must set the lease above plausible database interruption, renewal latency, and scheduling stalls.

Schema state is also fail closed. PostgreSQL schema version 2 stores an explicit singleton version and validates the expected layout. Unknown, absent, or forward versions are refused. Migration requires drained workers and reservations, a backup, the recognized migration, and explicit reopen. Live restore includes node and arbiter tables; audit-only recovery need not reopen admission. Thus gate rejection records a refusal without execution; lost reserve or settle acknowledgment is recovered idempotently or reported as uncertain; an operation exception releases its reservation without a completed node; lease loss is durable when persistence succeeds; and an unknown schema never opens implicitly.

\subsection{Deployment controls}

Production hardening separates compatibility coverage from systems evidence. Package integration tests exercise PostgreSQL 14 through 18 and Python 3.10 through 3.13, while the confirmatory concurrency study uses one pinned PostgreSQL 17.6 instance with synchronous commit enabled. The wider matrix checks schema creation, version refusal, recognized migration, reserve and settle idempotence, and ordinary reconnect behavior. E2 supplies the narrower but adversarial evidence for process races, delayed operations, lost acknowledgments, database restart, and lease loss.

The operator procedure is deliberately conservative. Before migration, workers are drained and active reservations must be reconciled. A database backup is taken before the recognized forward migration; an unknown or newer version is never modified automatically. Restoring service requires both node and arbiter tables, followed by explicit schema validation. Seal custody is backed up independently from the governed store. These steps do not create high availability, but they prevent an application restart from silently converting an unknown database or uncertain settlement into accepted work.

\subsection{Action admission}

Actions are registered with a name, version, JSON schema, side-effect declaration, and specification digest. The registry digest commits to the sorted set of action specifications and can be attached to a run. Admission rejects unknown actions, version mismatches, invalid arguments, policy violations, or missing confirmation. Schema annotations such as descriptions, titles, and defaults remain in the specification digest even when they do not change validation. This permits discovery through MCP and other tool registries without making a moving external registry authoritative during replay.

These checks are structural. A policy callback can inspect declared tool metadata and arguments, but Pollard does not classify arbitrary content as safe or unsafe and does not replace a semantic guardrail language such as AgentSpec \cite{wang2025agentspec}. The specification digest instead makes the exact action contract part of the recorded operating context.

\subsection{Integrity and telemetry}

Verification checks node identity, result digests, parent existence, and deterministic child order before a subtree is replayed, sealed, exported, or imported. Export writes a self-contained ordered manifest and its rolling seal; import verifies the full artifact before inserting any node. These checks expose changes within the sealed material under the stated hash assumption. They do not prevent a privileged operator from replacing both the store and a locally held seal, and metadata excluded from the seal is not protected.

The reference custody sink therefore writes the seal digest, store and root identifiers, sequence, time, algorithm, and signer identity to a separate SQLite database. It refuses to share the Pollard store file. The sink is a reference boundary, not a key-management system: signing, independent storage, access control, retention, and custody review remain operator responsibilities.

Telemetry exports operation kind, timing, outcome, and selected digests without making trace or span identifiers part of semantic identity. Payload and result content need not enter telemetry. This preserves compatibility with observability conventions such as OpenTelemetry \cite{opentelemetry2026semconv} while keeping operational correlation distinct from the integrity boundary.

\subsection{Framework integration}

Adapters remain intentionally thin. LangGraph functions, OpenAI Agents custom models and function tools, pydantic-ai function models, and MCP clients call the same Pollard model and tool primitives; Pollard does not translate their control graphs into its own scheduler. Framework events are recorded separately from model and tool boundaries. This distinction prevents a three-node graph from being misreported as three provider calls and allows an independent proxy to expose work that bypasses the chosen boundary.

\section{Evaluation}

The evaluation asks whether one semantic identity can support admission, replay, and audit without taking ownership of the caller's agent loop. Deterministic studies ran from the signed tag \texttt{ictai-confirmatory-v1}. Hosted attempts used separate signed tags, an independent spend guard, immutable terminal records, and no model substitution. All prespecified terminal outcomes are retained in the analysis.

\subsection{Design and analysis}

Each study fixes its unit, factors, oracle, exclusions, and terminal states. E1 uses an independent identity and budget state model. E2 runs fresh PostgreSQL process groups with barrier schedules and connection, acknowledgment, restart, retry, and lease faults. E3 pairs recording and replay with an out-of-band call sentinel, exact state digest, ancestry divergences, result mutations, and an argument-only cache. E4 measures each cell in a fresh process. E6 mutates exported artifacts, native objects, log bytes, and relational rows while an independent seal or snapshot remains outside the target store. E7 compares manual wrapping with LangGraph, OpenAI Agents, pydantic-ai, and MCP paths while an external proxy counts model and tool boundaries.

E5 uses six pinned hosted model identifiers and 15 result-blind selected $\tau^3$ and AgentDojo tasks \cite{yao2024taubench,barres2026tau3fixes,debenedetti2024agentdojo}. Its 81 adjacent pairs compare unwrapped execution with Pollard recording under inactive gates. Provider retries and hosted tools are disabled. A shared 50,000-input and 4,000-output token ledger applies to each trial, and the independent guard reserves worst-case cost before generation. The authorized total is 8.00 USD, with a 7.00 USD hard stop and 1.00 USD unavailable to the study.

Binary zero-event results use one-sided exact 95 percent bounds at the independent-unit level. E5 utility differences use task-cluster resampling with 10,000 fixed-seed replicates. Its prespecified analysis assigns zero utility to terminal errors rather than excluding them. Latency and cost are paired only within complete pairs. Timing maxima in E4 establish an observed resource envelope, not an asymptotic claim. Table~\ref{tab:evaluation-matrix} summarizes scope and completion.

\begin{table*}[t]
\caption{Frozen studies, independent units, and terminal coverage.}
\label{tab:evaluation-matrix}
\centering
\scriptsize
\setlength{\tabcolsep}{3pt}
\begin{tabular}{p{0.045\textwidth}p{0.18\textwidth}p{0.40\textwidth}p{0.29\textwidth}}
\toprule
Study & Unit & Principal oracle and contrasts & Terminal evidence \\
\midrule
E1 & State or seeded trace & Independent identity, budget, replay, and seal model; ten ablations & \EOneReplicas{} exact replicas, \EOneTransitions{} transitions each \\
E2 & Process group or injected fault & Conservation; transactional reservation versus three weaker counters; lease boundary & \ETwoGroups{} groups, \ETwoAttempts{} attempts, 20 fault cases \\
E3 & Record-replay pair & External-call sentinel, state digest, divergence oracle, cache aliasing & \EThreeCells{} cells, \EThreeCustody{} external seal publications \\
E4 & Fresh process & Store, depth, branch, payload reuse, interning, lifecycle operation & \EFourCells{} cells, \EFourProcesses{} measured processes \\
E5 & Reset task pair & Native scorer, token ledger, spend guard, inactive recording & E5-A: \EHostedATrials{} terminal trials; E5-B: \EHostedBTrials{} terminal trials \\
E6 & Trace and mutation & Independent verifier or externally held native snapshot & 1,680 terminal cases, \ESixControls{} clean controls \\
E7 & Manual-framework pair & Out-of-band event proxy, exact state and text parity & \ESevenPairs{} pairs, \ESevenTrees{} verified trees \\
\bottomrule
\end{tabular}
\end{table*}

\subsection{Identity, admission, and replay}

Both E1 replicas reproduced the same claim digest. Each executed \EOneTransitions{} seeded transitions after bounded exploration, with \EOneViolations{} full-model violations. Every one of the \EOneAblations{} required ablations exposed its preregistered witness. The result supports identity convergence, ancestry separation, refusal evidence, exact-budget invariants, strict replay, and seal sensitivity within the modeled bounds.

E2 completed \ETwoGroups{} one-host process groups and \ETwoAttempts{} admission attempts. None of the \ETwoRetainedGroups{} exact Pollard groups whose completing owner retained its lease violated conservation; the one-sided group-level upper bound is \ETwoUpper{}. Each weaker exact comparator produced barrier-synchronized overshoot. Estimate ratios behaved as predicted rather than as exact caps. In \ETwoLate{} suspended-owner cases, expired capacity was reused and the original owner later completed, producing visible overshoot outside the retained-lease theorem. Lost reserve acknowledgment, duplicate settlement, reconnect, restart, and ambiguous settlement retries reached terminal audited outcomes.

Across \EThreeCells{} E3 cells, strict replay entered the wrapped operation \EThreeStrictCalls{} times in \EThreeStrictObservations{} observations. The divergence oracle localized all \EThreeDivergences{} changed paths, all \EThreeTamperRejections{} stored-result mutations were rejected, and hybrid replay produced no unsafe reuse. In contrast, the argument-only cache created \EThreeUnsafeAliases{} unsafe aliases. This is the direct cost of omitting ancestry from a reuse key.

Table~\ref{tab:adverse-controls} collects the mechanism controls that distinguish a working guarantee from a passing workload. Each control changed only the named premise or instrumentation boundary. The three weaker exact counters each accumulated \ETwoWeakOvershoot{} units of overshoot. Boundary ablations preserved application behavior but made their omissions visible to the independent proxy. These outcomes show that the independent oracles discriminate the claimed mechanism from weaker alternatives.

\begin{table}[t]
\caption{Preregistered mechanism controls and observed witnesses.}
\label{tab:adverse-controls}
\centering
\scriptsize
\setlength{\tabcolsep}{3pt}
\begin{tabular}{p{0.39\columnwidth}p{0.53\columnwidth}}
\toprule
Control & Observed witness \\
\midrule
Post-hoc, local-lock, or settled-only counter & \ETwoWeakOvershoot{} aggregate overshoot units each \\
Expired reservation owner & \ETwoLate{} late completions outside exact theorem \\
Argument-only result cache & \EThreeUnsafeAliases{} ancestry-unsafe aliases \\
Model-only boundary & \ESevenModelOnlyMissing{} declared tool events absent \\
Tool-only boundary & \ESevenToolOnlyMissing{} declared model events absent \\
Permitted metadata edits & \ESixFalseAlarms{} alarms in \ESixControls{} controls \\
\bottomrule
\end{tabular}
\end{table}

\subsection{Resource, integrity, and portability results}

All \EFourCells{} E4 cells completed their five measured starts after warmup, totaling \EFourProcesses{} fresh processes. The largest observed process took \EFourMaxSeconds{} seconds and reached \EFourMaxRSSMiB{} MiB. In the four targeted high-reuse interning contrasts, enabled-to-disabled bytes per node had median ratio \EFourInterningRatio{}, but the ratios ranged from 0.252 on SQLite to 0.959 on PostgreSQL. We therefore report the individual contrasts and do not claim one store-independent saving or unwrapped runtime overhead.

E6 detected \ESixDetections{} of \ESixMutations{} semantic mutations and raised \ESixFalseAlarms{} semantic alarms on \ESixControls{} permitted metadata controls. It separately classified \ESixRepresentation{} representation-only changes, \ESixConstraint{} constraint-prevented mutations, and \ESixInapplicable{} representation-inapplicable cases. Whole-store deletion and replacement were judged against externally retained digests; local verification alone is not credited for that threat.

Every E7 pair matched the manual reference in final state and output. All \ESevenTrees{} Pollard trees verified. Full-boundary cells captured \ESevenFullCaptured{} of \ESevenFullDeclared{} declared events. Model-only and tool-only ablations omitted exactly the boundaries they did not wrap. Thus portability is supported for the tested releases and public interfaces, not for arbitrary background calls or framework extensions.

\subsection{Hosted validation and integration findings}

E5-A exercised all \EHostedATrials{} frozen trials, yielding \EHostedACompleted{} completions and \EHostedAErrors{} visible compatibility errors. Strict-schema tool limits, SDK token-count serialization, unresolved schema references, and weak exception capture identified concrete portability requirements for the hosted adapter. Retaining every terminal pair yielded 11 unwrapped and 10 inactive-Pollard utility successes, one unwrapped-only discordance, and mean paired difference \EHostedAUtilityDelta{} with task-cluster interval [\EHostedAUtilityCILow{}, \EHostedAUtilityCIHigh{}]. All 16 completed Pollard trials produced verified trees. Because harness-provider compatibility determined most terminal outcomes, the interval is reported for completeness and not interpreted as utility equivalence.

A separately signed corrective attempt incorporated the identified adapter fixes and passed exact token-count requests for all six largest frozen tool payloads without generation. E5-B then produced \EHostedBTrials{} terminal trials, or \EHostedBPairs{} complete pairs: \EHostedBCompleted{} completed and \EHostedBErrors{} returned explicit terminal errors. Each condition had five utility successes, with no discordance among these pairs, and all eight completed Pollard trials verified. Two Gemini 2.5 Pro trials reported model unavailability after successful token-count preflight, while two Gemini 2.5 Flash-Lite trials returned parallel tool calls outside the frozen single-call policy. These observations distinguish request-shape validation from live generation compatibility.

Guard reconciliation found that an E5-A child process had continued after its foreground terminal was interrupted, which made the initial E5-B freeze understate prior settlement. E5-B was stopped, the active reservation was charged at the full ambiguous ceiling, and process liveness plus guard state were reconciled. Conservative settlement across both attempts is \$\EHostedConservativeSpend{}, below the 7.00 USD hard stop, with no active reservation and the 1.00 USD reserve untouched. The remaining \$\EHostedRemainingGuard{} cannot fund the full frozen corrective schedule, and reducing the matrix after observing results is prohibited. This closeout preserves the preregistered budget and matrix rules. E5 therefore contributes end-to-end integration and portability evidence; it does not estimate benchmark-wide effectiveness.

\subsection{Interpretation}

The evidence supports three mechanism claims and supplies bounded external-validation evidence. First, ancestry is operational rather than decorative: removing it from the cache key produced unsafe reuse, while full strict replay returned verified bytes without a call. Second, shared admission is safe under the theorem's exact-meter and retained-lease premises. E2 preserved that invariant and distinguished the cases in which a premise no longer holds. Third, seal verification detects the declared semantic changes when the final digest has custody outside the attacked store. These conclusions rely on independent sentinels, models, snapshots, or proxies rather than on Pollard accepting its own records.

The hosted study addresses a different question: whether frozen provider stacks can carry the same boundary into live task execution. Every completed Pollard trial produced a valid tree, and E5-B observed no utility discordance across its ten complete pairs. Compatibility outcomes dominate E5-A, and E5-B is a result-dependent prefix rather than the frozen matrix. The result is therefore a verified end-to-end integration path plus a concrete portability boundary, not a utility-equivalence estimate and not a negative result about Pollard's deterministic mechanisms.

The hosted observations also yield operational requirements. Metadata access and token counting are necessary but insufficient provider gates. A model can accept token counting and reject generation, and schema dialects can differ on references, strict-tool counts, object closure, and parallel calls. A production deployment should validate its exact request shape, treat provider model identifiers as time-bound dependencies, preserve raw error responses, and supervise the actual child process rather than only its terminal. These controls establish provider portability before Pollard receives a completed boundary to govern.

\subsection{Reproducibility boundary}

Run manifests record package and paper commits, environment and database versions, model identifiers, request settings, task and schema digests, limits, terminal reasons, and settlements. Deterministic aggregates link to private raw inputs through a digest ledger. Hosted closeout records manifest-set digests and the external guard reconciliation without publishing licensed prompts or provider transcripts. Reproduction can audit every reported aggregate, but independent hosted reruns would be new stochastic experiments rather than exact replays.

The artifact keeps executable analysis separate from raw custody. Versioned summaries and TeX macros are regenerated from private manifest directories whose file-set digests are published. Protocol validators reject missing terminal states, stale source digests, unsupported model substitutions, and unsettled guard records. The deterministic suite reports an aggregate digest over its claim projections, while timing remains outside exact digest equality. This split permits exact audit of correctness claims without pretending that wall-clock measurements or hosted sampling are deterministic.

\section{Related Work}

\textbf{Resource governance.} Agent Contracts formalizes multi-dimensional constraints, temporal bounds, success criteria, and conservation under delegation \cite{ye2026contracts}. Token Budgets catalogs production incidents and studies an affine-typed Rust mitigation that makes several delegation errors compile-time failures \cite{khan2026budgets}. LiteLLM offers gateway-level budgets and rate limits for users and teams \cite{litellm2026budgets}. Pollard does not claim that preflight accounting is new. Its focus is the coupling of admission and settlement to the same ancestry-bound record used for replay.

\textbf{Runtime policy enforcement.} AgentSpec defines a language for runtime constraints and pre-action mediation \cite{wang2025agentspec}. A trace based assurance framework combines contracts, fault injection, deterministic replay, and governance over message action traces \cite{paduraru2026assurance}. Pollard contributes a concrete semantic identity and transactional store protocol. Its structural action registry is narrower than a general safety policy language.

\textbf{Replay, checkpoints, and caching.} LangGraph persists graph checkpoints and can fork from earlier state, while downstream nodes reexecute during time travel \cite{langchain2026timetravel}. AgentCheck records MCP tool responses and uses replay to reproduce injected failures \cite{mazumder2026agentcheck}. ToolCaching studies admission and eviction for repeated tool requests \cite{zhai2026toolcaching}. Pollard gives up some cross-context cache reuse in exchange for ancestry separation and strict nonexecution at a semantic boundary selected by the caller.

\textbf{Provenance and audit.} W3C PROV provides a general model of entities, activities, agents, and derivation \cite{moreau2013prov}. AiiDA records workflow provenance and supports durable, high throughput scientific computation \cite{huber2020aiida}. Recent agent research surveys typed execution graphs, evidence relations, and recovery requirements \cite{wang2026traces}. Lifecycle audit trails connect technical and governance records through tamper evident ledgers \cite{ojewale2026audit}. Pollard addresses step level operational provenance. It does not establish claim level evidence or organizational accountability by itself.

\textbf{Positioning.} These systems often identify different objects: a policy action, checkpoint, transport response, tool request, provenance activity, or accounting principal. Pollard does not replace them. It supplies a semantic step identifier that can be attached to their records when the application intends the same causal operation. The evaluation therefore avoids feature-count comparisons. Instead, it tests the consequences of the identity choice directly: ancestry changes must separate results, strict replay must avoid a call, reservation state must survive retry, and an omitted integration boundary must be observable. This framing also limits the novelty claim. Content addressing, leases, policy callbacks, checkpoints, and hash chains are established techniques; the contribution is their executable coupling at a caller-selected agent boundary and the failure model required to make that coupling auditable.

\section{Limitations and Threats to Validity}

\textbf{Semantic scope.} Pollard replays recorded results, not a hosted model's sampling process. Strict replay establishes nonexecution and stored-byte integrity; it cannot prove that the original service was correct or recreate changed external state. The caller also chooses the boundary. Work that bypasses an adapter is neither governed nor recorded.

\textbf{Resource assumptions.} Exact safety requires serializable reservation updates, exact estimates, and a lease held through settlement. Approximate meters inherit estimation error. An expired owner can finish after replacement work, as E2 demonstrated. During a store partition, Pollard refuses admission or reports uncertainty; it does not provide consensus, provider-side cancellation, or recovery of unreported usage. E2 used one physical host, so its results do not cover multi-host clock behavior, network partitions, or production contention.

\textbf{Integrity and measurement.} Content addressing evidences change under a hash assumption. A privileged adversary can replace a store and a colocated seal unless the final digest has independent custody. The reference sink separates files but does not provide key management, transparency, or independent administration. E4 reports one machine's absolute envelope and selected storage contrasts. Mutation and clean-control counts cover the declared corpus, not all faults or adversaries.

\textbf{External validity.} E7 tests selected framework releases and public interfaces with a small deterministic graph. E5 uses 15 tasks, six model identifiers, and a strict single-call policy. It exercised the full E5-A schedule and ten corrective pairs across three provider families, but compatibility constraints limited the usable paired sample. Model availability also changed between metadata and generation requests. These observations bound the hosted utility inference; they do not alter the deterministic mechanism results or establish benchmark-wide equivalence, model rankings, or small utility effects.

\textbf{Researcher control.} Signed tags, result-blind task selection, immutable terminal records, digest ledgers, fixed terminal treatment, and explicit closeout reduce selective reporting. They do not remove workload selection bias or substitute for an independent reproduction. E5 also shows that a signed manifest must be paired with process-liveness and external-guard reconciliation before a later freeze.

\section{Conclusion}

Pollard tests a systems proposition: admission, execution, refusal, replay, and audit become mutually checkable when they share one ancestry-bound semantic identity. The caller retains the agent loop and selects model and tool boundaries; Pollard supplies the identity, transactional reserve-settle lifecycle, strict stored-result replay, structural action admission, and rolling evidence.

The signed deterministic suite supports those mechanisms under their stated premises. State exploration preserved all invariants while every ablation exposed a witness. Retained-lease PostgreSQL groups stayed within exact limits, strict replay made no external call, all declared semantic mutations were detected without a control alarm, and 360 framework pairs preserved manual behavior while verifying 720 trees. Mechanism controls confirmed that these outcomes depend on transactional reservation, retained leases, ancestry-aware reuse, complete boundary placement, and external seal custody.

Hosted validation extended the evaluation into live provider paths. Every completed Pollard trial yielded a verified tree, and the corrective sample showed no observed utility discordance. Provider compatibility and fixed-budget rules left the planned benchmark comparison incomplete, so we make no utility-equivalence claim. The main result remains narrow and testable: within a correctly instrumented boundary, retained lease, and externally anchored seal, one semantic tree can connect governance decisions to replayable execution evidence.

\bibliographystyle{IEEEtran}
\bibliography{refs}

@article{ye2026contracts,
  author        = {Qing Ye and Jing Tan},
  title         = {Agent Contracts: A Formal Framework for Resource-Bounded Autonomous {AI} Systems},
  journal       = {arXiv preprint arXiv:2601.08815},
  year          = {2026},
  eprint        = {2601.08815},
  archiveprefix = {arXiv},
  primaryclass  = {cs.MA},
  doi           = {10.48550/arXiv.2601.08815},
  url           = {https://arxiv.org/abs/2601.08815}
}

@article{khan2026budgets,
  author        = {Sajjad Khan},
  title         = {Token Budgets: An Empirical Catalog of 63 {LLM}-Agent Budget-Overrun Incidents, with an Affine-Typed Rust Mitigation as a Case Study},
  journal       = {arXiv preprint arXiv:2606.04056},
  year          = {2026},
  eprint        = {2606.04056},
  archiveprefix = {arXiv},
  primaryclass  = {cs.SE},
  doi           = {10.48550/arXiv.2606.04056},
  url           = {https://arxiv.org/abs/2606.04056}
}

@article{wang2025agentspec,
  author        = {Haoyu Wang and Christopher M. Poskitt and Jun Sun},
  title         = {{AgentSpec}: Customizable Runtime Enforcement for Safe and Reliable {LLM} Agents},
  journal       = {arXiv preprint arXiv:2503.18666},
  year          = {2025},
  eprint        = {2503.18666},
  archiveprefix = {arXiv},
  primaryclass  = {cs.AI},
  doi           = {10.48550/arXiv.2503.18666},
  note          = {Accepted by the 48th IEEE/ACM International Conference on Software Engineering, 2026},
  url           = {https://arxiv.org/abs/2503.18666}
}

@article{mazumder2026agentcheck,
  author        = {Aritra Mazumder and Nusrat jahan Lia},
  title         = {{AgentCheck}: A Reproduce-Intervene-Mitigate Workbench for {LLM} Agents over {MCP}},
  journal       = {arXiv preprint arXiv:2607.11098},
  year          = {2026},
  eprint        = {2607.11098},
  archiveprefix = {arXiv},
  primaryclass  = {cs.SE},
  doi           = {10.48550/arXiv.2607.11098},
  url           = {https://arxiv.org/abs/2607.11098}
}

@article{zhai2026toolcaching,
  author        = {Yi Zhai and Dian Shen and Junzhou Luo and Bin Yang},
  title         = {{ToolCaching}: Towards Efficient Caching for {LLM} Tool-calling},
  journal       = {arXiv preprint arXiv:2601.15335},
  year          = {2026},
  eprint        = {2601.15335},
  archiveprefix = {arXiv},
  primaryclass  = {cs.SE},
  doi           = {10.48550/arXiv.2601.15335},
  url           = {https://arxiv.org/abs/2601.15335}
}

@article{wang2026traces,
  author        = {Yiqi Wang and Jiaqi Zhang and Taotao Cai and Zirui Liu and Qingqiang Sun and Zequn Sun and Zhangkai Wu and Manqing Dong and Mingkai Zheng and Xuefei Yin and Yanming Zhu},
  title         = {From Agent Traces to Trust: A Survey of Evidence Tracing and Execution Provenance in {LLM} Agents},
  journal       = {arXiv preprint arXiv:2606.04990},
  year          = {2026},
  eprint        = {2606.04990},
  archiveprefix = {arXiv},
  primaryclass  = {cs.CR},
  doi           = {10.48550/arXiv.2606.04990},
  url           = {https://arxiv.org/abs/2606.04990}
}

@article{paduraru2026assurance,
  author        = {Ciprian Paduraru and Petru-Liviu Bouruc and Alin Stefanescu},
  title         = {A Trace-Based Assurance Framework for Agentic {AI} Orchestration: Contracts, Testing, and Governance},
  journal       = {arXiv preprint arXiv:2603.18096},
  year          = {2026},
  eprint        = {2603.18096},
  archiveprefix = {arXiv},
  primaryclass  = {cs.MA},
  doi           = {10.48550/arXiv.2603.18096},
  url           = {https://arxiv.org/abs/2603.18096}
}

@article{ojewale2026audit,
  author        = {Victor Ojewale and Harini Suresh and Suresh Venkatasubramanian},
  title         = {Audit Trails for Accountability in Large Language Models},
  journal       = {arXiv preprint arXiv:2601.20727},
  year          = {2026},
  eprint        = {2601.20727},
  archiveprefix = {arXiv},
  primaryclass  = {cs.CY},
  doi           = {10.48550/arXiv.2601.20727},
  url           = {https://arxiv.org/abs/2601.20727}
}

@article{huber2020aiida,
  author  = {Sebastiaan P. Huber and Spyros Zoupanos and Martin Uhrin and Leopold Talirz and Leonid Kahle and Rico H{\"a}uselmann and Dominik Gresch and Tiziano M{\"u}ller and Aliaksandr V. Yakutovich and Casper W. Andersen and Francisco F. Ramirez and Carl S. Adorf and Fernando Gargiulo and Snehal Kumbhar and Elsa Passaro and Conrad Johnston and Andrius Merkys and Andrea Cepellotti and Nicolas Mounet and Nicola Marzari and Boris Kozinsky and Giovanni Pizzi},
  title   = {{AiiDA} 1.0, a Scalable Computational Infrastructure for Automated Reproducible Workflows and Data Provenance},
  journal = {Scientific Data},
  volume  = {7},
  pages   = {300},
  year    = {2020},
  doi     = {10.1038/s41597-020-00638-4},
  url     = {https://www.nature.com/articles/s41597-020-00638-4}
}

@techreport{moreau2013prov,
  author      = {{World Wide Web Consortium}},
  editor      = {Luc Moreau and Paolo Missier},
  title       = {{PROV-DM}: The {PROV} Data Model},
  institution = {World Wide Web Consortium},
  type        = {{W3C} Recommendation},
  year        = {2013},
  month       = apr,
  url         = {https://www.w3.org/TR/prov-dm/}
}

@misc{langchain2026timetravel,
  author       = {{LangChain}},
  title        = {Use time-travel},
  year         = {2026},
  howpublished = {LangGraph documentation},
  note         = {Accessed 2026-07-21},
  url          = {https://docs.langchain.com/oss/python/langgraph/use-time-travel}
}

@misc{litellm2026budgets,
  author       = {{LiteLLM}},
  title        = {Budgets, Rate Limits},
  year         = {2026},
  howpublished = {LiteLLM documentation},
  note         = {Accessed 2026-07-21},
  url          = {https://docs.litellm.ai/docs/proxy/users}
}

@misc{opentelemetry2026semconv,
  author       = {{OpenTelemetry Authors}},
  title        = {{OpenTelemetry} Semantic Conventions 1.43.0},
  year         = {2026},
  howpublished = {OpenTelemetry specification},
  note         = {Accessed 2026-07-21},
  url          = {https://opentelemetry.io/docs/specs/semconv/}
}

@article{yao2024taubench,
  author        = {Shunyu Yao and Noah Shinn and Pedram Razavi and Karthik Narasimhan},
  title         = {{$\tau$-bench}: A Benchmark for Tool-Agent-User Interaction in Real-World Domains},
  journal       = {arXiv preprint arXiv:2406.12045},
  year          = {2024},
  eprint        = {2406.12045},
  archiveprefix = {arXiv},
  primaryclass  = {cs.AI},
  doi           = {10.48550/arXiv.2406.12045},
  url           = {https://arxiv.org/abs/2406.12045}
}

@misc{barres2026tau3fixes,
  author       = {Victor Barres and Ben Shi},
  title        = {{$\tau^3$-Bench}: Fixing Airline + Retail},
  year         = {2026},
  month        = feb,
  howpublished = {Sierra Research},
  note         = {Accessed 2026-07-21},
  url          = {https://taubench.com/blog/tau3-task-fixes.html}
}

@article{debenedetti2024agentdojo,
  author        = {Edoardo Debenedetti and Jie Zhang and Mislav Balunovi{\'c} and Luca Beurer-Kellner and Marc Fischer and Florian Tram{\`e}r},
  title         = {{AgentDojo}: A Dynamic Environment to Evaluate Prompt Injection Attacks and Defenses for {LLM} Agents},
  journal       = {arXiv preprint arXiv:2406.13352},
  year          = {2024},
  eprint        = {2406.13352},
  archiveprefix = {arXiv},
  primaryclass  = {cs.CR},
  doi           = {10.48550/arXiv.2406.13352},
  url           = {https://arxiv.org/abs/2406.13352}
}

\end{document}